\documentclass[conference]{IEEEtran}
\IEEEoverridecommandlockouts
\usepackage{flushend}
\usepackage{cite}
\usepackage{amsmath,amssymb,amsfonts}
\usepackage{algorithmic}
\usepackage{algorithm}
\usepackage{graphicx}
\usepackage{textcomp}
\usepackage{xcolor}
\usepackage{siunitx}
\usepackage{booktabs}
\usepackage{subcaption}
\usepackage{overpic}
\usepackage{adjustbox}
\usepackage{tabularx}
\DeclareMathOperator*{\argmin}{argmin} % thin space, limits underneath in displays
\def\BibTeX{{\rm B\kern-.05em{\sc i\kern-.025em b}\kern-.08em
    T\kern-.1667em\lower.7ex\hbox{E}\kern-.125emX}}
\begin{document}

% \title{HIMT: Hierarchical Feature Disentanglement-driven Intermediate Modality Translation for MRI-Ultrasound Prostate Registration
% \title{ACMT: Anatomically coherent Modality Translation for MRI-Ultrasound Prostate Registration
\title{Modality Translation and Registration of MR and Ultrasound Images Using Diffusion Models
% {\footnotesize \textsuperscript{*}Note: Sub-titles are not captured for https://ieeexplore.ieee.org  and
}

\author{
Xudong Ma\textsuperscript{1},
Nantheera Anantrasirichai\textsuperscript{1},
Stefanos Bolomytis\textsuperscript{2}, 
Alin Achim\textsuperscript{1}\\
\textsuperscript{1}Visual Information Laboratory, University of Bristol, Bristol, the United Kingdom \\
\textsuperscript{2}Southmead Hospital, North Bristol NHS Trust, Bristol, the United Kingdom \\
\{xudong.ma, n.anantrasirichai, alin.achim\}@bristol.ac.uk, stefanos.bolomytis@nbt.nhs.uk
}

% \author{\IEEEauthorblockN{Xudong Ma}
% \IEEEauthorblockA{\textit{ University of Bristol}\\
% Bristol, UK \\
% xudong.ma@bristol.ac.uk}
% \and
% \IEEEauthorblockN{Nantheera Anantrasirichai}
% \IEEEauthorblockA{\textit{University of Bristol}\\
% Bristol, UK \\
% n.anantrasirichai@bristol.ac.uk}
% \and
% \IEEEauthorblockN{Stefanos Bolomytis}
% \IEEEauthorblockA{\textit{Southmead Hospital}\\
% Bristol, UK \\
% stefanos.bolomytis@nbt.nhs.uk}
% \and
% \IEEEauthorblockN{Alin Achim}
% \IEEEauthorblockA{\textit{University of Bristol}\\
% Bristol, UK \\
% alin.achim@bristol.ac.uk}
% \and
% \and

\maketitle

\begin{abstract}
Multimodal MR-US registration is critical for prostate cancer diagnosis. However, this task remains challenging due to significant modality discrepancies. Existing methods often fail to align critical boundaries while being overly sensitive to irrelevant details. To address this, we propose an anatomically coherent modality translation (ACMT) network based on a hierarchical feature disentanglement design. We leverage shallow-layer features for texture consistency and deep-layer features for boundary preservation. Unlike conventional modality translation methods that convert one modality into another, our ACMT introduces the customized design of an intermediate pseudo modality. Both MR and US images are translated toward this intermediate domain, effectively addressing the bottlenecks faced by traditional translation methods in the downstream registration task. Experiments demonstrate that our method mitigates modality-specific discrepancies while preserving crucial anatomical boundaries for accurate registration. Quantitative evaluations show superior modality similarity compared to state-of-the-art modality translation methods. Furthermore, downstream registration experiments confirm that our translated images achieve the best alignment performance, highlighting the robustness of our framework for multi-modal prostate image registration.
\end{abstract}


\section{Introduction}
Prostate cancer is a leading cause of death among men worldwide~\cite{bratt2024population}, with accurate diagnosis relying heavily on advanced imaging techniques. Magnetic Resonance Imaging (MRI) and Ultrasound (US) are widely used in prostate cancer imaging, each offering unique advantages: MRI provides high soft-tissue contrast for lesion detection, while US enables real-time guidance for biopsies. However, their significant differences in anatomical representation pose a major challenge for multi-modal registration, which is essential for combining their complementary strengths.

To address modality discrepancies, some methods employ image segmentation, aligning segmented regions to avoid direct cross-modal registration challenges \cite{chen2021mr,jiang2023segmentation}. However, these approaches require extensive annotated data, which are labor-intensive and scarce, limiting their practical application. To overcome these limitations, we previously proposed a Partial Modality Translation (PMT) approach \cite{ma2024pmt}, which established an intermediate modality to bridge MRI and US domains. While PMT improved texture similarity, it does not truly achieve customized modality translation and retains excessive details irrelevant to registration. This oversight left room for improvement in the registration results.

Building on this foundation, we propose a novel hierarchical feature disentanglement method for anatomically coherent modality translation  (ACMT). Through a customized intermediate modality design, ACMT logically fulfills the goal that PMT failed to achieve. The ACMT advances PMT by addressing two key limitations: (1) explicitly preserving boundary information while maintaining texture consistency, and (2) relaxing the requirement for photorealism so as to reduce the preservation of unnecessary and overly detailed textures within the prostate. Specifically, we leverage the distinct characteristics of shallow and deep features in convolutional networks. Shallow-layer features, which primarily capture low-level texture information, are processed with larger convolutional kernels. This design is motivated by the fact that larger kernels can effectively model broader texture patterns, which are essential for achieving cross-modal consistency \cite{youwang2024paint}. In contrast, deep-layer features, which are more sensitive to high-level structural details such as shapes, are processed with smaller convolutional kernels. Smaller kernels are better suited for capturing fine-grained details and precise anatomical structures, making them ideal for boundary preservation \cite{purwono2022understanding}. Additionally, we enhance boundary extraction using Sobel filtering to further improve structural alignment \cite{ranjan2023edge}. In addition, by removing the adversarial loss used in PMT, we further reduce the emphasis on photorealism, as the intermediate modality does not require highly realistic details. This design ensures that the output retains only the boundary information most relevant for registration, while promoting texture consistency and suppressing modality-specific anatomical detail discrepancies. As a result, the difficulty of cross-modal registration is significantly reduced.

In the subsequent sections of this paper, we will provide a comprehensive explanation of the mathematical foundations and network architecture of the proposed method in Section 2. Section 3 will present both objective and subjective analyses of the results, demonstrating the effectiveness of our approach. Finally, we will conclude with a summary of our contributions in the last section.

\section{Proposed Method}
\begin{figure*}
    \centering
    \includegraphics[width=\textwidth]{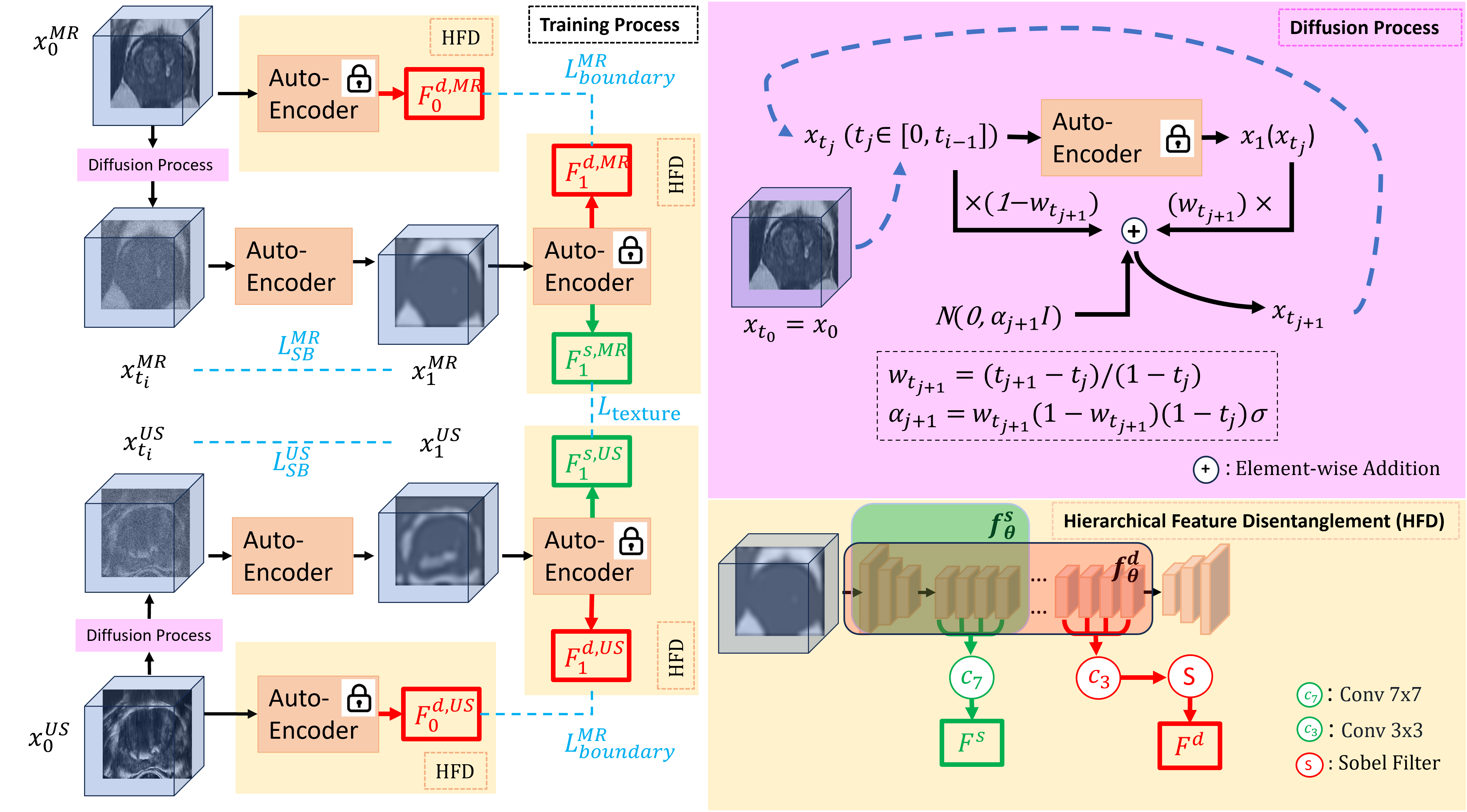}
    \caption{Hierarchical Feature Disentanglement framework based on a Diffusion Model}
    \label{fig:network}
\end{figure*}

In this section, we first introduce the theoretical foundation of our method, which is based on a Diffusion Schrödinger Bridge. Following this, we provide a detailed explanation of the overall workflow of the proposed ACMT framework, as illustrated in Fig.~\ref{fig:network}.
% \subsection{Network Architecture}

\subsection{Schrödinger Bridge based Diffusion Model}
We design this network by leveraging a Schrödinger Bridge based (SB) diffusion framework \cite{kim2024unpaired}. It is inspired by optimal transport (OT) theory \cite{wang2021deep,leonard2014survey}. Formally, the SB problem seeks to find the optimal stochastic process that transforms a source distribution $P_0$ to a target distribution $P_1$ through intermediate distributions $P_t$ over time t, as defined by:
\begin{equation}
P^{SB} = \{\argmin_{P_t} D_{KL}(P_t \lVert W^\sigma)\} \quad  with\quad t\sim[0,1],
\end{equation}
% \begin{equation}
% P^{SB} = \argmin_{P \in \mathcal{P}(P_0, P_1)} D_{KL}(P \lVert W^\sigma),
% \end{equation}
where $W^\sigma$ denotes the Wiener measure with variance 
$\sigma$. This formulation aims to minimize the Kullback-Leibler (KL) divergence between the process distribution $P_t$ and the reference measure $W^\sigma$ at each timestep. The collection of optimal distributions $\{P_t^{SB}\}$ constitutes the Schrödinger Bridge connecting $P_0^{SB}$ and $P_1^{SB}$. 
Although SB is theoretically a continuous process, Conditional Flow Matching (CFM) formulation allows us to address it in a discretized manner \cite{kim2024unpaired,tong2023conditional}. The CFM formulation establishes that for any two distributions 
\( P_{t_m}^{SB} \) and \( P_{t_n}^{SB} \) in the SB where 
\([t_m,t_n] \subseteq [0,1]\), the intermediate distribution at time 
\(t \in [t_m,t_n]\) follows a Gaussian distribution:
\begin{multline}
p(X_t|X_{t_m},X_{t_n}) = \mathcal{N}\Big(X_t \Big| w_tX_{t_n}+(1-w_t)X_{t_m}, \\
w_t(1-w_t)\sigma(t_n-t_m)\textbf{\textit{I}}\Big),
\label{eq:conditional}
\end{multline}
where $X_t \sim P_t^{SB}$, $X_{t_m} \sim P_{t_m}^{SB}$, $X_{t_n} \sim P_{t_n}^{SB}$ and $w_t=\frac{t-t_m}{t_n-t_m}$.
Furthermore, the joint distribution $P_{t_m t_n}^{SB}$ between any two timesteps can be obtained through an entropy-regularized OT problem:
\begin{multline}
P_{t_m t_n}^{SB} = \argmin_{P_{t_m,t_n}} E_{(X_{t_m},X_{t_n})}[\lVert X_{t_m}-X_{t_n}\lVert^2] \\
- 2\sigma(t_n-t_m)H(X_{t_m},X_{t_n}),
\label{eq:optjoint}
\end{multline}
where \( H \) denotes the entropy function. This formulation enables the determination of the optimal terminal distribution \( P_{t_n}^{SB} \) given any initial distribution \( P_{t_m}^{SB} \). Subsequently, using Equation (\ref{eq:conditional}), we can compute any intermediate state \( P_{t}^{SB} \) along the Schrödinger Bridge between the \(P_{t_m}^{SB} \) and \(P_{t_n}^{SB} \).

\subsection{Diffusion Process via CFM formulation}

Since the training of our model relies on inputs generated via a diffusion process, we first provide a comprehensive explanation of the diffusion process that forms the foundation of our framework. This diffusion process applies of our modality translation network. However, it is used in evaluation mode with all parameters frozen. To be more specific, we hypothesize the existence of an ideal intermediate modality that focuses primarily on boundary information while disregarding texture details. Through careful network design and loss function formulation, our framework learns to transform both MR and US images into this target modality. This process is achieved by constructing Schr\"{o}dinger Bridge from either MR  \(P_0^{MR}\) distribution or US distribution \(P_0^{US}\) to a single intermediate modality \(P_1\).

As illustrated in the Diffusion Process (purple block) in Fig.~\ref{fig:network}, let \(x_{t_j} \in P_{t_j}\) represent an intermediate state along the Schrödinger Bridge at time \(t_j \in [0,1]\). Our network learns to map \(x_{t_j}\) to the terminal state \(x_1 \in P_1\). By using the CFM formulation as shown in Equation \ref{eq:conditional}, the next state \(x_{t_{j+1}}\) could be computed as:

\begin{equation}
x_{t_{j+1}} = w_{t_{j+1}} x_1 + (1 - w_{t_{j+1}}) x_{t_j} + \mathcal{N}(0, \alpha_{j+1} I),
\label{eq:diffusion}
\end{equation}

where \(w_{t_{j+1}} = \frac{t_{j+1} - t_j}{1 - t_j}\) is the interpolation weight balancing \(x_1\) and \(x_{t_j}\), \(\alpha_{j+1} = w_{t_{j+1}} (1 - w_{t_{j+1}}) (1 - t_j) \sigma\) controls the noise magnitude, and \(\mathcal{N}(0, \alpha_{j+1} I)\) adds Gaussian noise scaled by \(\alpha_{j+1}\).

% This formulation combines deterministic interpolation with adaptive noise injection, enabling flexible and controlled transitions toward the terminal state.

This iterative process begins with \(x_{t_0} = x_0\), the source MR or US image, and progressively transforms it towards \(x_{t_i}\).

\subsection{Training Process}

The training process begins with inputs generated by a diffusion process. As shown in the Training Process (left part of Fig.\ref{fig:network}), during training, we implement the following procedure for each MR-US image pair:

\begin{enumerate}
    \item Randomly select $t_i$ from the predefined time step pool $\{t_0, t_1, t_2, \ldots, t_T\}$, where each $t_i \in [0, 1]$.
    \item Using the network in evaluation mode (with all parameters frozen): generate \(x_{t_i}\) through the diffusion process as described in the purple block of Fig. \ref{fig:network}
    
    \item Switch the network to training mode:
    \begin{itemize}
        \item Compute the transformation from \(x_{t_i}\) to \(x_1\)
        \item Calculate loss 
        \item Update network parameters through backpropagation
    \end{itemize}
\end{enumerate}

Through this training strategy, the network learns to transform any intermediate state \(x_{t_i}\) to the target modality \(x_1\), effectively capturing the mapping along the entire Schr\"{o}dinger Bridge. By breaking down the transformation into iterative steps, it ensures gradual refinement of the target modality while preserving critical anatomical information. This approach is particularly advantageous for medical imaging, where the complex relationship between source and target modalities requires robust solutions.

\subsection{Loss Functions}

Recent studies have demonstrated the remarkable capability of diffusion models in extracting discriminative features from images \cite{preechakul2022diffusion,yang2023diffusion}. Building upon this observation and the well-established understanding that shallow layers of neural networks are more sensitive to texture information while deeper layers capture boundary details \cite{purwono2022understanding}, we design a hierarchical feature disentanglement loss to achieve both texture consistency and boundary preservation in our modality translation framework.

As shown in the Hierarchical Feature Disentanglement part of Fig. \ref{fig:network}, let \(f_\theta^s\) and \(f_\theta^d\) denote the shallow and deep feature extraction functions of our network, respectively. Given an input image \(x_0\) (either MR or US), the shallow and deep features are extracted as:
\begin{equation}
\mathbf{F}_0^s = f_\theta^s(x_0), \quad \mathbf{F}_0^d = f_\theta^d(x_0).
\end{equation}
Similarly, for the translated image \(x_1\) in the intermediate modality \(P_1\), the corresponding features are:
\begin{equation}
\mathbf{F}_1^s = f_\theta^s(x_1), \quad \mathbf{F}_1^d = f_\theta^d(x_1).
\end{equation}

\subsubsection{Texture Consistency Loss}
To ensure texture similarity between the translated MR and US images in the intermediate domain, we further process the shallow features \(\mathbf{F}_1^{s,MR}\) and \(\mathbf{F}_1^{s,US}\) using a large convolutional kernel (\(7 \times 7\)) because larger kernels are known to be more effective at capturing general, global features such as texture patterns due to their wider receptive fields. The texture consistency loss is formulated as:
\begin{equation}
\mathcal{L}_{\text{texture}} = \left\| \mathcal{C}_{7\times7}(\mathbf{F}_1^{s,MR}) - \mathcal{C}_{7\times7}(\mathbf{F}_1^{s,US}) \right\|_2^2,
\end{equation}
where \(\mathcal{C}_{7\times7}(\cdot)\) represents the \(7 \times 7\) convolution operation, and \(\|\cdot\|_2^2\) denotes the L2 norm.

\subsubsection{Boundary Preservation Loss}
To preserve the anatomical boundaries of the original images, we process the deep features \(\mathbf{F}_0^d\) and \(\mathbf{F}_1^d\) using a smaller convolutional kernel (\(3 \times 3\)) followed by a Sobel filter. This design choice is motivated by the fact that smaller kernels are particularly effective at extracting fine-grained features, such as edges and boundaries, due to their ability to focus on localized regions and capture high-frequency details. The boundary preservation loss for MR and US images is defined as:
\begin{equation}
\mathcal{L}_{\text{boundary}}^{MR} = \left\| \mathcal{S}(\mathcal{C}_{3\times3}(\mathbf{F}_1^{d,MR})) - \mathcal{S}(\mathcal{C}_{3\times3}(\mathbf{F}_0^{d,MR})) \right\|_1,
\end{equation}
\begin{equation}
\mathcal{L}_{\text{boundary}}^{US} = \left\| \mathcal{S}(\mathcal{C}_{3\times3}(\mathbf{F}_1^{d,US})) - \mathcal{S}(\mathcal{C}_{3\times3}(\mathbf{F}_0^{d,US})) \right\|_1,
\end{equation}
where \(\mathcal{C}_{3\times3}(\cdot)\) denotes the \(3 \times 3\) convolution operation, \(\mathcal{S}(\cdot)\) represents the Sobel filter, and \(\|\cdot\|_1\) is the L1 norm. The total boundary preservation loss is:
\begin{equation}
\mathcal{L}_{\text{boundary}} =  \frac{1}{2} \left(
\mathcal{L}_{\text{boundary}}^{MR} + \mathcal{L}_{\text{boundary}}^{US} \right).
\end{equation}

\subsubsection{SB Loss}
To ensure that the translation process follows the optimal transport path defined by the Schr\"{o}dinger Bridge, we introduce the SB constraint loss based on the joint distribution \(P_{t_i,1}^{SB}\) following the theory in equation \ref{eq:optjoint}. For MR and US images, the SB loss is defined as:
\begin{equation}
\begin{aligned}
\mathcal{L}_{\text{SB}}^{MR}(\theta_i,t_i) &= \mathbb{E}_{(x_{t_i}^{MR},x_1^{MR})} \left[ \| x_{t_i}^{MR} - x_1^{MR} \|^2 \right] \\
&\quad - 2\sigma(1-t_i)H(x_{t_i}^{MR},x_1^{MR}),
\end{aligned}
\end{equation}
\begin{equation}
\begin{aligned}
\mathcal{L}_{\text{SB}}^{US}(\theta_i,t_i) &= \mathbb{E}_{(x_{t_i}^{US},x_1^{US})} \left[ \| x_{t_i}^{US} - x_1^{US} \|^2 \right] \\
&\quad - 2\sigma(1-t_i)H(x_{t_i}^{US},x_1^{US}).
\end{aligned}
\end{equation}
where \(x_1^{MR} = f_{\theta_i}(x_1^{MR}|x_{t_i}^{MR})\) and \(x_1^{US} = f_{\theta_i}(x_1^{US}|x_{t_i}^{US})\) are the terminal states predicted by the network. The total SB loss is then:
\begin{equation}
\mathcal{L}_{\text{SB}} = \frac{1}{2} \left( \mathcal{L}_{\text{SB}}^{MR} + \mathcal{L}_{\text{SB}}^{US} \right).
\end{equation}

% \subsubsection{Adversarial Loss}
% Finally, to enhance the visual quality of the generated images, we employ a generative adversarial network (GAN) with a discriminator \(D\) and a generator \(G\). The adversarial loss is designed to ensure that the generated images \(x_1^{MR}\) and \(x_1^{US}\) are visually realistic and consistent with the target modality. Specifically, the discriminator \(D\) is trained to classify real US images (\(x_0^{US}\)) as true and both real MR images (\(x_0^{MR}\)) and generated images (\(x_1^{MR}\) and \(x_1^{US}\)) as false. This design choice is motivated by the fact that US images in our dataset exhibit higher local resolution and clearer boundary information compared to MR images, making them a more suitable reference for generating high-quality outputs:
% \begin{equation}
%   \begin{aligned}
%   \mathcal{L}_{\text{adv}}^D &= \mathbb{E}_{x_0^{US}}[\log D(x_0^{US})] + \mathbb{E}_{x_0^{MR}}[\log (1 - D(x_0^{MR}))] \\
%   &\quad + \mathbb{E}_{x_1^{MR}}[\log (1 - D(x_1^{MR}))] + \mathbb{E}_{x_1^{US}}[\log (1 - D(x_1^{US}))].
%   \end{aligned}
% \end{equation}

% The generator \(G\) (our translation network) is trained to fool the discriminator by making the generated images \(x_1^{MR}\) and \(x_1^{US}\) indistinguishable from real US images:
% \begin{equation}
% \mathcal{L}_{\text{adv}}^G = \mathbb{E}_{x_1^{MR}}[\log D(x_1^{MR})] + \mathbb{E}_{x_1^{US}}[\log D(x_1^{US})].
% \end{equation}

The overall loss function combines the texture consistency loss, boundary preservation loss, and the SB loss:
\begin{equation}
\mathcal{L}_{\text{total}} = \lambda_{\text{texture}} \mathcal{L}_{\text{texture}} + \lambda_{\text{boundary}} \mathcal{L}_{\text{boundary}} + \lambda_{\text{SB}} \mathcal{L}_{\text{SB}} 
% + \lambda_{\text{adv}} \mathcal{L}_{\text{adv}}^G
\end{equation}

The weighting coefficients \(\lambda_{\text{texture}}\), \(\lambda_{\text{boundary}}\), and \(\lambda_{\text{SB}}\) are carefully tuned to balance the contributions of each loss component, ensuring that the network simultaneously achieves texture consistency, boundary preservation, and optimal transport.

\begin{figure*}
    \centering

    % 第一行标签
    \begin{subfigure}{0.12\textwidth}
        \centering \scriptsize \textbf{US}
    \end{subfigure}
    \begin{subfigure}{0.12\textwidth}
        \centering \scriptsize \textbf{MR}
    \end{subfigure}
    \begin{subfigure}{0.12\textwidth}
        \centering \scriptsize \textbf{UNSB-MR}
    \end{subfigure}  
    \begin{subfigure}{0.12\textwidth}
        \centering \scriptsize \textbf{PMT-US}
    \end{subfigure}
    \begin{subfigure}{0.12\textwidth}
        \centering \scriptsize \textbf{PMT-MR}
    \end{subfigure}
    \begin{subfigure}{0.12\textwidth}
        \centering \scriptsize \textbf{ACMT-US}
    \end{subfigure}
    \begin{subfigure}{0.12\textwidth}
        \centering \scriptsize \textbf{ACMT-MR}
    \end{subfigure}
    
    \vspace{0.5em} % 标题和图像间距
    
    % 第一行图片（Patient 1）
    \begin{subfigure}{0.12\textwidth}
        \centering
        \includegraphics[width=\textwidth]{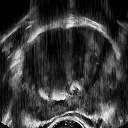}
    \end{subfigure}
    \begin{subfigure}{0.12\textwidth}
        \centering
        \includegraphics[width=\textwidth]{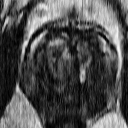}
    \end{subfigure}
    \begin{subfigure}{0.12\textwidth}
        \centering
        \includegraphics[width=\textwidth]{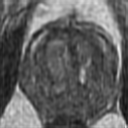}
    \end{subfigure}  
    \begin{subfigure}{0.12\textwidth}
        \centering
        \includegraphics[width=\textwidth]{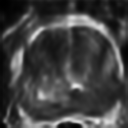}
    \end{subfigure}
    \begin{subfigure}{0.12\textwidth}
        \centering
        \includegraphics[width=\textwidth]{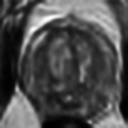}
    \end{subfigure}
    \begin{subfigure}{0.12\textwidth}
        \centering
        \includegraphics[width=\textwidth]{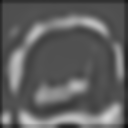}
    \end{subfigure}
    \begin{subfigure}{0.12\textwidth}
        \centering
        \includegraphics[width=\textwidth]{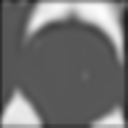}
    \end{subfigure}
    
    \vspace{0.1em} 
    
    % 第二行图片（Patient 2）
    \begin{subfigure}{0.12\textwidth}
        \centering
        \includegraphics[width=\textwidth]{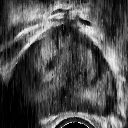}
    \end{subfigure}
    \begin{subfigure}{0.12\textwidth}
        \centering
        \includegraphics[width=\textwidth]{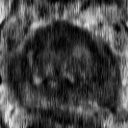}
    \end{subfigure}
    \begin{subfigure}{0.12\textwidth}
        \centering
        \includegraphics[width=\textwidth]{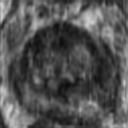}
    \end{subfigure}  
    \begin{subfigure}{0.12\textwidth}
        \centering
        \includegraphics[width=\textwidth]{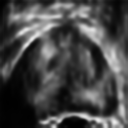}
    \end{subfigure}
    \begin{subfigure}{0.12\textwidth}
        \centering
        \includegraphics[width=\textwidth]{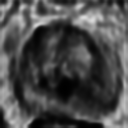}
    \end{subfigure}
    \begin{subfigure}{0.12\textwidth}
        \centering
        \includegraphics[width=\textwidth]{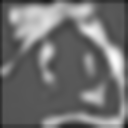}
    \end{subfigure}
    \begin{subfigure}{0.12\textwidth}
        \centering
        \includegraphics[width=\textwidth]{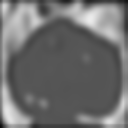}
    \end{subfigure}

    \caption{Modality translation results for two patients (two rows), showing original US and MR images, UNSB translation, and intermediate translation from PMT and ACMT.}
    \label{fig:translation}
\end{figure*}

\begin{figure*}
    \centering

    % 第一行：Patient 1
    \begin{subfigure}{0.16\textwidth}
        \centering
        \textbf{US}\\[0.3em]
        \includegraphics[width=\textwidth]{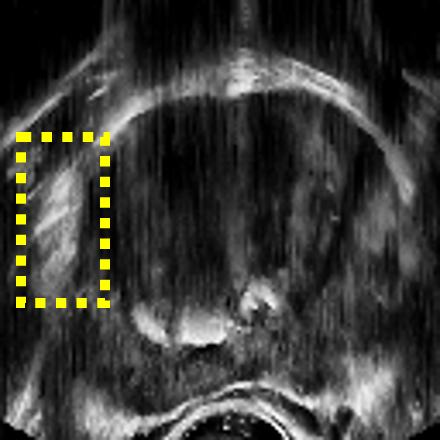}
    \end{subfigure}
    \begin{subfigure}{0.16\textwidth}
        \centering
        \textbf{MR}\\[0.3em]
        \includegraphics[width=\textwidth]{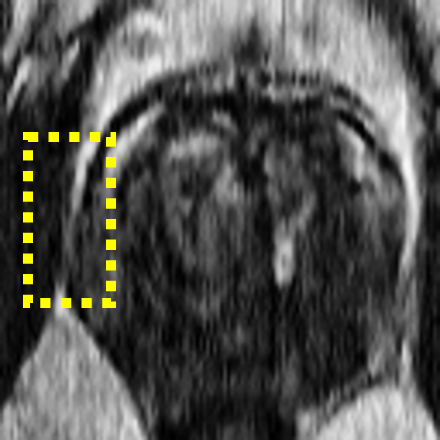}
    \end{subfigure}
    \begin{subfigure}{0.16\textwidth}
        \centering
        \textbf{original+FSDiffReg}\\[0.3em]
        \includegraphics[width=\textwidth]{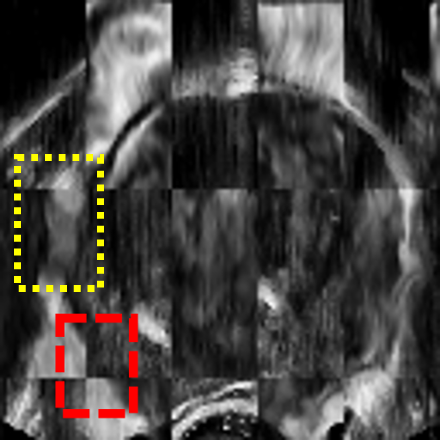}
    \end{subfigure}
    \begin{subfigure}{0.16\textwidth}
        \centering
        \textbf{UNSB+FSDiffReg}\\[0.3em]
        \includegraphics[width=\textwidth]{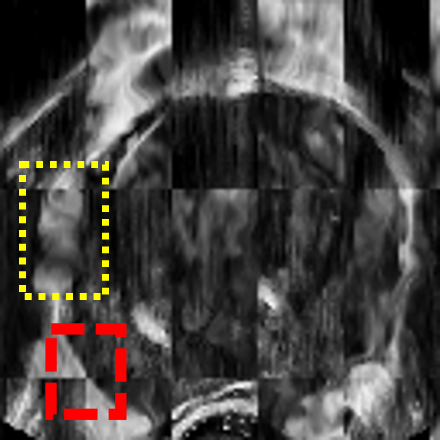}
    \end{subfigure}
    \begin{subfigure}{0.16\textwidth}
        \centering
        \textbf{PMT+FSDiffReg}\\[0.3em]
        \includegraphics[width=\textwidth]{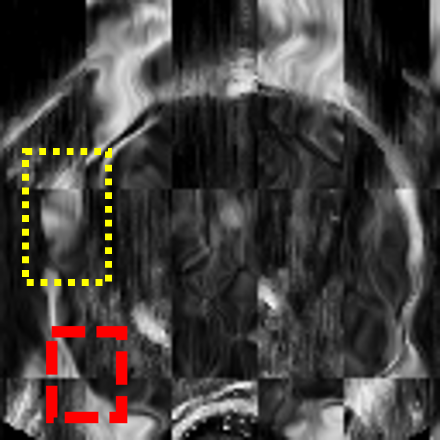}
    \end{subfigure}
    \begin{subfigure}{0.16\textwidth}
        \centering
        \textbf{ACMT+FSDiffReg}\\[0.3em]
        \includegraphics[width=\textwidth]{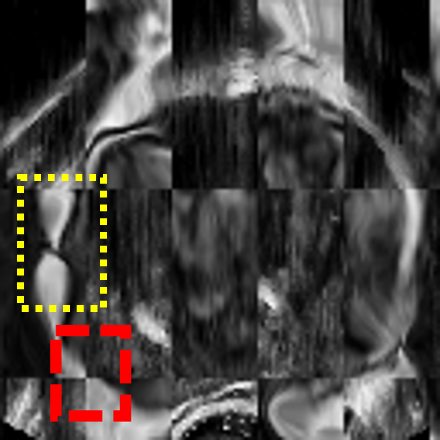}
    \end{subfigure}
    \\[0.1em] % 行间距
    % 第二行：Patient 2
    \begin{subfigure}{0.16\textwidth}
        \centering
        \includegraphics[width=\textwidth]{images/Patient_17_0.png}
    \end{subfigure}
    \begin{subfigure}{0.16\textwidth}
        \centering
        \includegraphics[width=\textwidth]{images/Patient_17_1.png}
    \end{subfigure}
    \begin{subfigure}{0.16\textwidth}
        \centering
        \includegraphics[width=\textwidth]{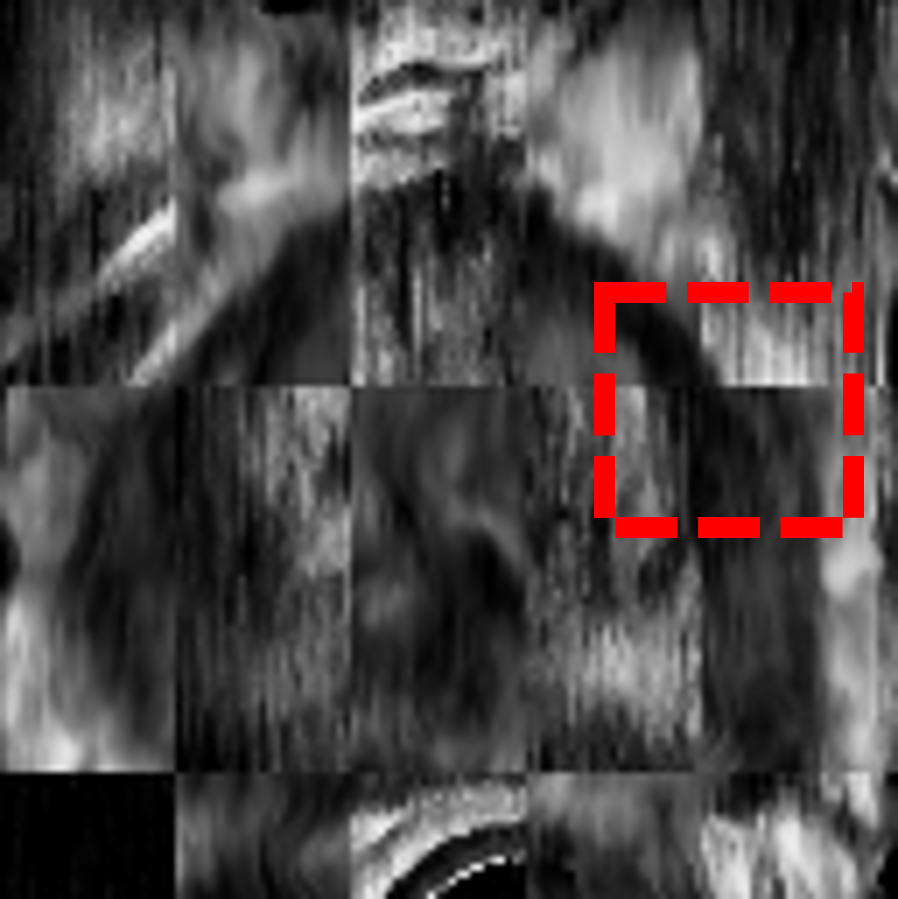}
    \end{subfigure}
    \begin{subfigure}{0.16\textwidth}
        \centering
        \includegraphics[width=\textwidth]{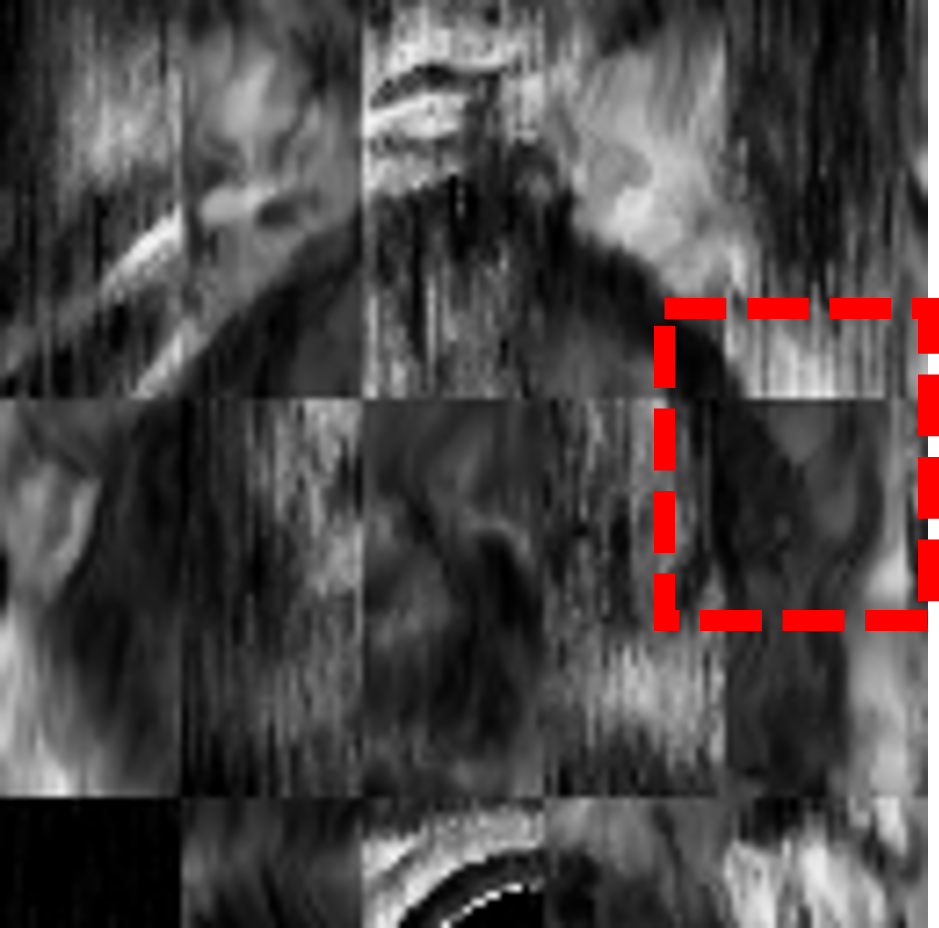}
    \end{subfigure}
    \begin{subfigure}{0.16\textwidth}
        \centering
        \includegraphics[width=\textwidth]{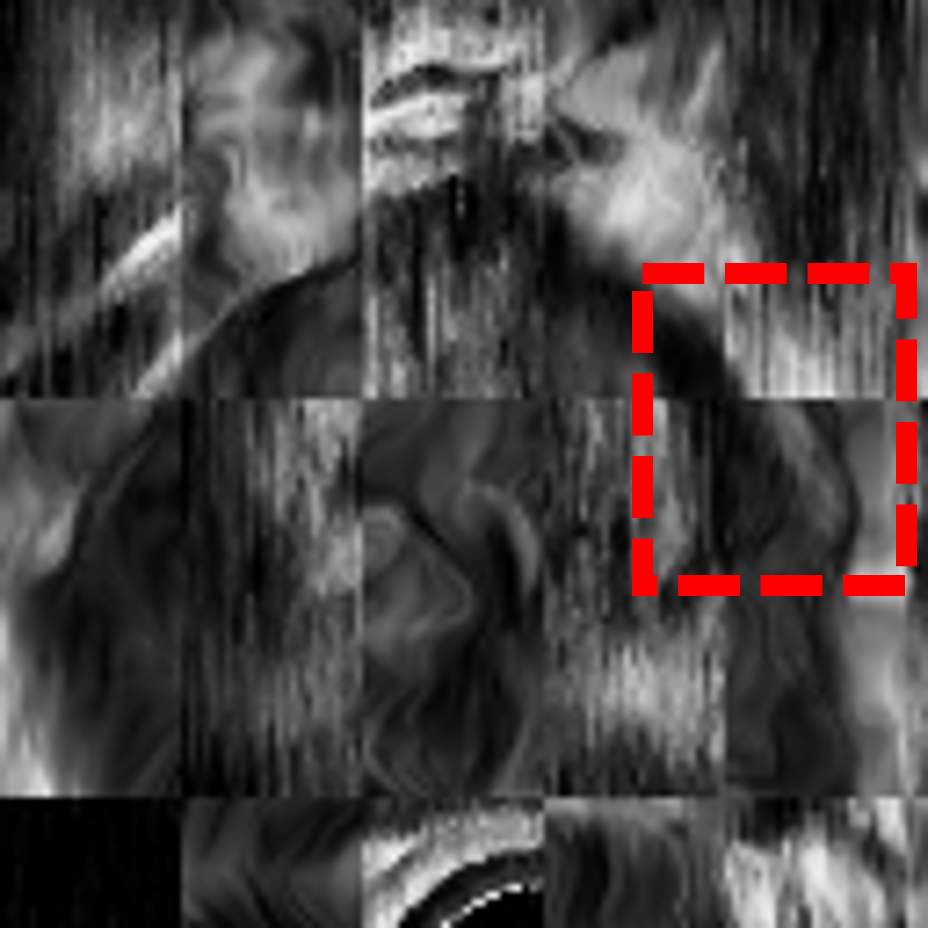}
    \end{subfigure}
    \begin{subfigure}{0.16\textwidth}
        \centering
        \includegraphics[width=\textwidth]{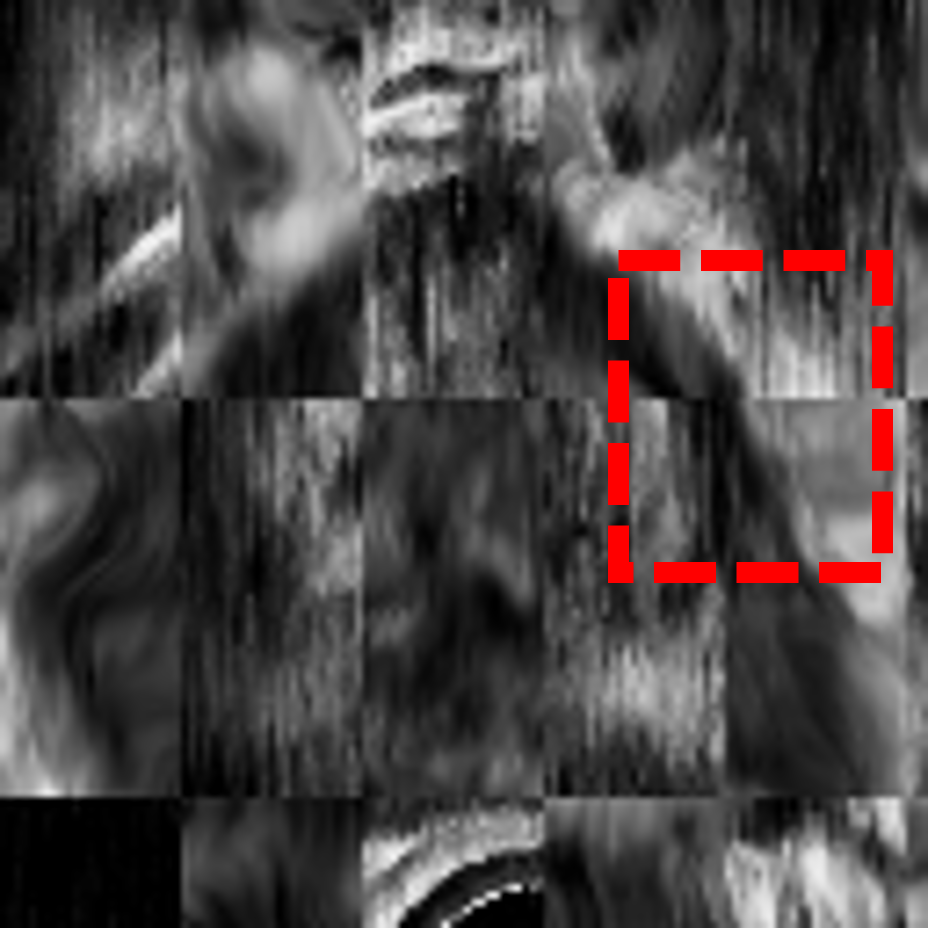}
    \end{subfigure}

    \caption{Registration results for two patients. Each row corresponds to one patient and displays the original US and MR images, followed by registration results using FSDiffReg applied to the original MR-US pair, as well as to images translated by UNSB, PMT, and our ACMT methods. Results are shown in a chessboard layout for visual comparison.}
    \label{fig:registration}
\end{figure*}

% As shown in Figure~\ref{fig:network}, our network architecture consists of:
% \begin{itemize}
%     \item Two 3D convolutional downsampling blocks
%     \item Nine ResNet blocks for feature extraction
%     \item Two convolutional upsampling layers 
% \end{itemize}
% This architecture was designed through careful consideration of computational efficiency and experimental performance, effectively balancing the network's capacity with available GPU resources.

\section{Results and Evaluation}
\subsubsection{Dataset}  
3D MR and US paired images were provided by Southmead Hospital Bristol. The MR images are T2-weighted volumes, while the 3D US images are derived from biopsy procedures. To ensure consistency, all volumes were preprocessed to focus on the prostate region while retaining relevant surrounding structures. %The dataset consists of five 3D image pairs, and while all are used for analysis, two cases are presented in detail to illustrate typical outcomes.  
 Additionally, to enhance generalizability and robustness, we applied data augmentation techniques such as flipping and rotation. To balance computational feasibility and anatomical fidelity, all volumes were standardized to a resolution of 128$\times$128$\times$64 pixels. The dataset was partitioned into an 80\% training set and a 20\% test set, and all experiments employed cross-validation.

\begin{table}[t]
    \centering
    \caption{Quantitative evaluation of modality translation quality using FID and KID (lower is better).}
    \label{tab:fid_kid}
    \begin{tabularx}{\linewidth}{Xcc}  % X 表示自动调整列宽
        \toprule
        Method & FID $\downarrow$ (decrease by $\uparrow$) & KID $\downarrow$ (decrease by $\uparrow$)  \\
        \midrule
        Original & 404.88 & 0.56 \\
        UNSB & 377.92 (6.66\%) & 0.52 (7.14\%)  \\
        PMT  & 170.02 (58.01\%) & 0.11 (80.36\%) \\
        ACMT(Ours) & \textbf{138.01 (65.91\%)} & \textbf{0.09 (83.93\%)} \\
        \bottomrule
    \end{tabularx}
\end{table}

\begin{table}[t]
    \centering
    \caption{Quantitative evaluation of registration performance based on different modality translation methods, all registered using a consistent approach.}
    \label{tab:ACMT_registration}
    \begin{tabular}{lcccc}
        \toprule
        Method & DSC $\uparrow$ & IoU $\uparrow$ & ASD $\downarrow$\\
        \midrule
        % Original & 0.89 & 0.81 & 12.65 & {--}\\
        Original & 0.92 & 0.87 & 10.74  \\
        UNSB & 0.92 & 0.88 & 12.83  \\
        PMT & \textbf{0.95} & \textbf{0.91} & 9.18 \\
        ACMT (ours) & \textbf{0.95} & 0.90 & \textbf{6.82} \\ 
        \bottomrule
    \end{tabular}
\end{table}
 
\subsubsection{Evaluation}
We evaluated both the quality of modality translation and the corresponding registration performance using the same registration model, FSDiffReg~\cite{qin2023fsdiffreg}. We compared: (1) registration on original MR-US images, and (2) registration on modality-translated results from state-of-the-art (SOTA) approaches, including UNSB \cite{kim2024unpaired}, an ICLR 2024 modality translation method, and PMT \cite{ma2024pmt}, our previously proposed method that focuses on texture similarity.
% It is important to note that all modality translation results are solely used to generate the warping map, while the actual warping is applied to the original images. This design ensures that even if minor artifacts are introduced during modality translation, they have minimal impact on the final registration results.

To assess the quality of modality translation, we employed two widely-used metrics: FID and KID. For registration evaluation, our experts manually segmented the prostate on several key frames for each test case to create ground-truth masks. The generated deformation fields, based on different modality translation methods, were applied to warp these masks, and the results were scored with:
\begin{itemize}
  \item \textbf{Dice Similarity Coefficient (DSC)}: Defined as 
    \( \frac{2|X \cap Y|}{|X| + |Y|}\), this metric quantifies the
    volume overlap between the warped mask \(X\) and the ground-truth mask \(Y\).

  \item \textbf{Intersection-over-Union (IoU)}: Given by
    \( \frac{|X \cap Y|}{|X \cup Y|}\), IoU provides a stricter
    assessment of boundary alignment by measuring the ratio of the shared to the
    combined volume.

  \item \textbf{Average Surface Distance (ASD)}: This metric reports the mean
    Euclidean distance between the surfaces of the warped and ground-truth
    segmentations.
\end{itemize}

% Additionally, to quantify the smoothness of each deformation field, we computed its harmonic energy (HE), defined as the squared Frobenius norm of the Laplacian of the 3-D displacement field \(\phi = (\phi_1,\phi_2,\phi_3)\):
% \begin{equation}
% \mathcal{E}_{\text{HE}}
%   = \lVert \nabla^{2} \phi \rVert_{F}^{2}
%   = \sum_{i=1}^{3} \lVert \nabla^{2} \phi_{i} \rVert^{2}.
% \end{equation}
% A lower HE value indicates a smoother, and therefore more physically plausible transformation.

The experimental results demonstrate that our method achieves the best FID and KID scores, as shown in Table~\ref{tab:fid_kid}. Specifically, our approach reduces the FID by 65.91\% and the KID by 83.93\%, outperforming UNSB approximately 10 times and 12 times, respectively. Compared to our previously proposed PMT method, our current method shows further improvements in both metrics, highlighting its superior performance in modality translation. On the other hand, as shown in Table \ref{tab:ACMT_registration}, our ACMT framework attains the best overall registration results, leading all methods in DSC and ASD. While its IoU is only 0.01 below that of PMT, ACMT lowers ASD by 25.7 \% compared with the runner-up, demonstrating superior surface alignment. These findings confirm that ACMT produces more anatomically faithful intermediate representations and yields substantially higher registration accuracy than either UNSB or PMT-based variants.

Visually, we randomly selected two patients to illustrate the modality translation and image registration comparison. As shown in Fig.~\ref{fig:translation}, Our method maximizes the similarity within the prostate interior while preserving boundary information that is most relevant for registration. At the same time, it suppresses unnecessary internal texture details that could mislead the registration process. This advantage is further validated in the registration results shown in Fig.~\ref{fig:registration}. In particular, in the second row, within the red boxes, our method is the only one that achieves relatively smooth registration, while all other methods exhibit noticeable discontinuities. Similarly, in the first patient case (top row), although all modality translation approaches help improve the smoothness in the red-boxed region, the yellow boxes highlight that our method introduces the least over-deformation. This suggests that our estimated deformation field is the most realistic and anatomically plausible among all methods.

\section{Conclusion}
In this work, we propose a novel ACMT method for modality translation between MR and US images using a hierarchical feature disentanglement idea. We leverage shallow features to ensure texture consistency and deep features to preserve anatomical boundaries, resulting in anatomically coherent pseudo-representations. Our unsupervised framework achieves customized modality translation, effectively removing irrelevant information from source images that would otherwise hinder cross-modal registration. Experimental results show that our framework consistently outperforms SOAT methods in both modality translation and registration, achieving superior performance in both quantitative metrics and visual assessments.

%This approach not only ensures robust registration but also demonstrates potential for broader multi-modal applications.

% Future work will focus on extending our framework to other multi-modal registration tasks and exploring its applicability in real-time clinical settings. Additionally, we plan to investigate the integration of additional anatomical constraints to further improve registration accuracy and robustness.

\bibliographystyle{splncs04}
\bibliography{mybibliography}

\end{document}